\documentclass{article}

\usepackage{microtype}
\usepackage{graphicx}
\usepackage{subcaption}
\usepackage{booktabs} 
\usepackage{siunitx}
\usepackage[table]{xcolor}
\usepackage{enumitem}
\usepackage[most]{tcolorbox}
\usepackage{amsmath,amssymb}

\newcommand{\pubpartial}{$\triangle$}
\newcommand{\pubyes}{O}
\newcommand{\pubno}{X}

\DeclareMathOperator{\Ind}{Ind}
\tcbuselibrary{breakable,skins,listings}

\usepackage{hyperref}

\newlength{\panelgap}
\usepackage[preprint]{icml2026}

\usepackage{amsmath}
\usepackage{amssymb}
\usepackage{mathtools}
\usepackage{amsthm}

\tcbset{
  promptbox/.style={
    breakable, enhanced,
    colback=blue!05, colframe=blue!20,
    coltitle=black, fonttitle=\bfseries\color{black},
    sharp corners, boxrule=0.6pt,
    left=1.2mm,right=1.2mm,top=1mm,bottom=1mm
  }
}

\usepackage[capitalize,noabbrev]{cleveref}

\theoremstyle{plain}

\theoremstyle{definition}

\theoremstyle{remark}

\usepackage[textsize=tiny]{todonotes}

\newcommand{\benchmark}{\textsc{ExpertMath{} }}
\icmltitlerunning{Judging What We Cannot Solve: A Consequence-Based Approach for Oracle-Free Evaluation of Research-Level Math}
\begin{document}

\twocolumn[
  \icmltitle{Judging What We Cannot Solve: A Consequence-Based \\Approach for Oracle-Free Evaluation of Research-Level Math}



  \icmlsetsymbol{equal}{*}

  \begin{icmlauthorlist}
    \icmlauthor{Guijin Son}{snu,olar}
    \icmlauthor{Donghun Yang}{kisti}
    \icmlauthor{Hitesh Laxmichand Patel}{oracle} 
    \icmlauthor{Hyunwoo Ko}{olar} \\
    \icmlauthor{Amit Agarwal}{oracle}
    \icmlauthor{Sunghee Ahn}{snu}
    \icmlauthor{Kyong-Ha Lee}{kisti}
    \icmlauthor{Youngjae Yu}{snu}
  \end{icmlauthorlist}

  \icmlaffiliation{snu}{Seoul National University}
  \icmlaffiliation{olar}{OnelineAI}
  \icmlaffiliation{kisti}{KISTI}
  \icmlaffiliation{oracle}{ORACLE}

  \icmlcorrespondingauthor{Guijin Son}{guijin.son@snu.ac.kr}

  \vskip 0.3in
]



\printAffiliationsAndNotice{}  

\begin{abstract}
Recent progress in reasoning models suggests that generating plausible attempts for research-level mathematics may be within reach, but verification remains a bottleneck, consuming scarce expert time. We hypothesize that a meaningful solution should contain enough method-level information that, when applied to a neighborhood of related questions, it should yield better downstream performance than incorrect solutions. Building on this idea, we propose \textbf{Consequence-Based Utility}, an oracle-free evaluator that scores each candidate by testing its value as an in-context exemplar in solving related yet verifiable questions. Our approach is evaluated on an original set of research-level math problems each paired with one expert-written solution and nine LLM-generated solutions. Notably, Consequence-Based Utility consistently outperforms reward models, generative reward models, and LLM judges on ranking quality. Specifically, for GPT-OSS-120B it improves Acc@1 from 67.2 to 76.3 and AUC from 71.4 to 79.6, with similarly large AUC gains on GPT-OSS-20B (69.0 to 79.2). Furthermore, compared to LLM-Judges, it also exhibits a larger solver–evaluator gap, maintaining stronger correct–wrong separation even on instances the underlying solver often fails to solve.
\end{abstract}

\section{Introduction}

For a mathematical hypothesis to be accepted as scientific knowledge, it must undergo extensive review and validation. Yet many recent efforts to advance science with LLMs~\citep{gottweis2025towards} emphasize hypothesis generation~\citep{zhou2024hypothesis, radensky2024scideator} and experimental planning~\citep{goel2025training}, while giving comparatively less attention to rigorous validation. Accordingly, this step is largely dependent on either human experts~\citep{georgiev2025mathematical}, which are costly to scale, or LLM judges (including agentic systems)~\citep{lu2024ai,zhu2025safescientist, panigrahi2026heurekabench}, that are often unreliable~\citep{son2024llm, son2025ai} and biased~\citep{ye2024justice}. These limitations motivate the need for better methods for hypothesis validation.

\begin{figure}[t] 
\centering 
\includegraphics[width=\linewidth]{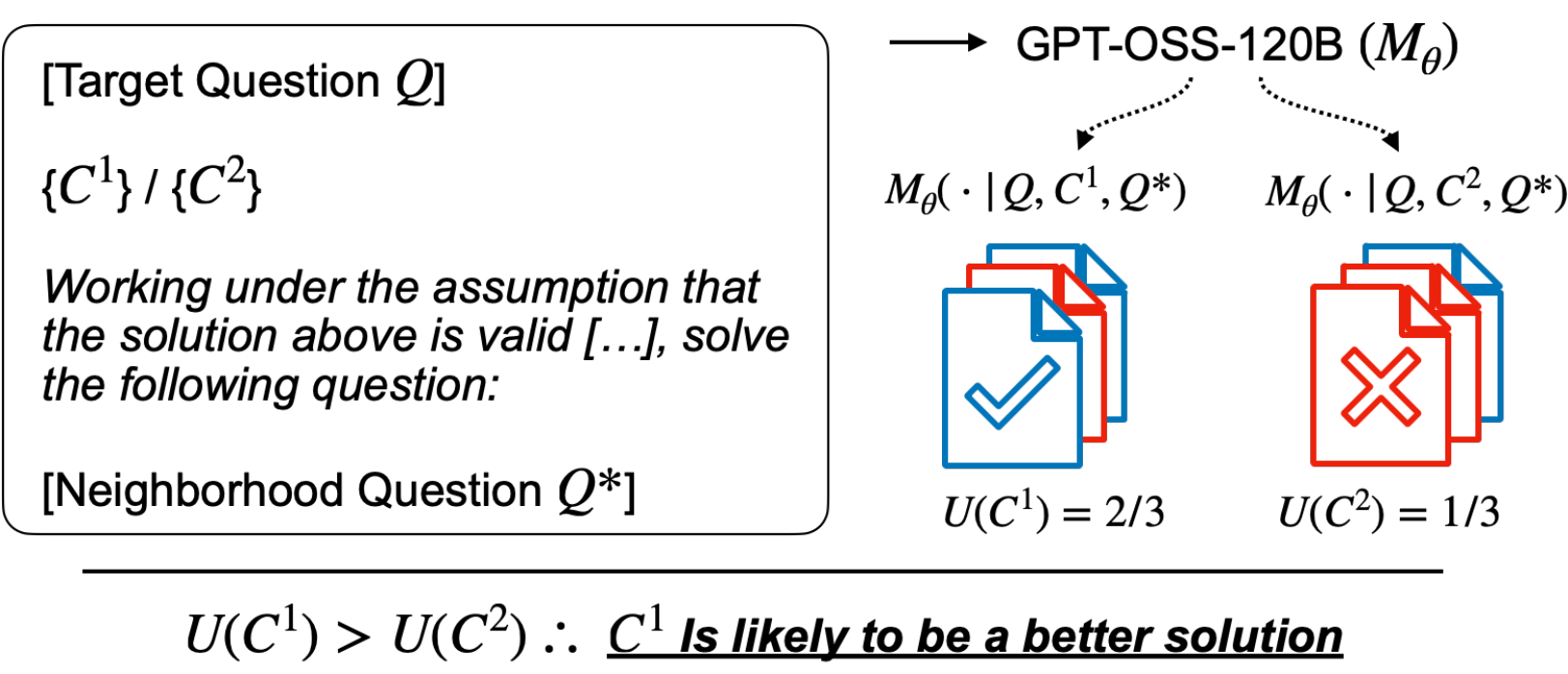} \caption{\textbf{Consequence-Based Utility for solution validation.} We use GPT-OSS-120B as the solver $M_\theta$ and score each candidate solution by its induced accuracy on neighborhood questions $Q^*$; $U(C^1) > U(C^2)$ suggests $C^1$ is more likely correct.
 }

\label{fig:cbu} 
\vspace{-3mm}
\end{figure}

In this work, we introduce \textbf{Consequence-Based Utility}, a novel approach to validate a set of candidate solutions without access to ground-truth answers. As shown in Figure~\ref{fig:cbu}, we prompt a solver $M_\theta$ with a research-level question $Q$ and $C^{(i)}$ as in-context exemplars. For each $C^{(i)}$, we measure the solver’s accuracy on a closely related neighborhood problem $Q^*$ and use the resulting accuracy as its utility score. Intuitively, a candidate that induces higher accuracy on $Q^*$, provides more helpful information for $Q$ and is therefore more likely to be correct. It should be noted that \emph{Consequence-Based Utility is designed to focus on research-level questions}, those remaining open to LLMs today. We therefore focus on genuine research-level questions that remain out of reach for today’s LLMs, and curate \benchmark consisting of 192 expert-written problems and 425 LLM-Generated questions. Half of the expert-written questions remain open to leading models (e.g., GPT-5 and Gemini-3-Pro). In this dataset, our method outperforms oracle-free baselines such as reward models, generative reward models, and LLM judges. For instance, as an LLM-Judge, GPT-OSS-120B achieves Acc@1 = 67.21 and AUC = 71.42; under Consequence-Based Utility, these increase to 76.27 and  79.63, respectively. Moreover, Consequence-Based Utility exhibits a larger solver--evaluator gap than LLM judges, preserving a stronger separation between correct and incorrect solutions even for questions the model fail to solve. This makes it particularly well-suited for evaluating research-level questions. Finally, our error analysis reveals that these gains arise from Consequence-Based Utility more reliably downranking solutions with incorrect reasoning, unjustified compression, or unjustified interpretation, and being less sensitive to stylistic cues and authority-like statements that are known to mislead LLM judges~\citep{ye2024justice, moon2025don}.

Our contributions are summarized as follows:

\begin{itemize}[topsep=0pt, itemsep=0.3em]
    \item We propose \textbf{Consequence-Based Utility}, an oracle-free method for validating candidate solutions via downstream performance on neighborhood questions.

    \item We release \benchmark a collection of $192$ expert-written research-level math problems with author solutions, along with $425$ LLM-generated problems.

    \item We show that \textsc{CBU} consistently outperforms oracle-free baselines (LLM-judges, reward models, and generative reward models), and identify judge failure modes that \textsc{CBU} reliably penalizes through error analysis.

    \item We provide a practitioner’s guide for \textsc{CBU}, including how to construct neighborhood questions and how many rollouts are needed for stable utility estimates.
\end{itemize}

\section{Preliminary and Related Works}

\subsection{Call for Oracle-Free Validation in Math}
Recent case studies indicate that LLMs can meaningfully assist professional mathematicians on genuine open or previously unsolved research problems. In late 2025, publicly documented human–LLM collaborations (i) established point convergence of Nesterov’s accelerated gradient method~\citep{jang2025point}, (ii) produced a finite counterexample to a “majority optimality” conjecture in non-interactive correlation distillation with erasures~\citep{ivanisvili2025counterexample}, and (iii) determined the sharp minimax-optimal error rate for robust density estimation under Wasserstein-bounded contamination~\citep{dobriban2025solving}. Despite the notable progress, however, these reports underscore that current models are high-variance generators rather than reliable autonomous theorem provers: \citet{jang2025point} reports that ChatGPT generated \emph{“numerous arguments, approximately 80\% of which were incorrect,”} \citet{dobriban2025solving} notes that GPT‑5 \emph{“glossed over details that sometimes took days of work to fill in,”} and \citet{schmitt2025extremal} observes that \emph{“Some models claimed false counterexamples.”} Consequently, progress still depends on professor-level triage. Experts must reject hallucinated proof attempts, repair missing steps, and translate ideas into checkable arguments before any result is safe to trust or share. These experiences motivate the need for oracle-free validation: scalable validation mechanisms that can filter and score candidate research outputs without requiring a scarce domain-expert oracle for each attempt.

\subsection{Existing Oracle-Free Validators.}
We model a \emph{candidate solution} as an object $C \in \mathcal{C}$ (e.g., a proof sketch, lemma chain, or an algorithmic construction) for a research question $Q \in \mathcal{Q}$. A generator LLM $M_\theta$ induces a conditional distribution over candidates,
\[
C^{(i)} \sim p_\theta(\,\cdot \mid Q), \qquad i=1,\dots,N.
\]
In an idealized setting, there exists a (typically unavailable) correctness oracle
\[
O(Q,C)\in\{0,1\},
\]
which returns $1$ iff $C$ is fully correct (and $0$ otherwise). ``Oracle-free validation'' replaces $O$ with a \emph{validator} $V$ that outputs a score used for selection or ranking:
\[
V:\mathcal{Q}\times\mathcal{C}\to\mathbb{R},\qquad 
\widehat{C}\;=\;\arg\max_{i\in[N]} V(Q,C^{(i)}).
\]
Below, we formalize three widely used validators: consistency voting~\citep{wang2022self}, reward models~\citep{ouyang2022training}, and LLM judges\citep{zheng2023judging}.

\paragraph{(1) Majority (consistency) voting.}
Majority voting assumes that each candidate $C$ deterministically induces a discrete prediction $A(C)\in\mathcal{A}$ (e.g., a numeric answer or yes/no). Given $N$ i.i.d. samples $C^{(1:N)}$ with induced answers $A^{(i)} := A(C^{(i)})$, the majority-vote answer is $\widehat{A}_{\mathrm{mv}} := \arg\max_{a\in\mathcal{A}} \sum_{i=1}^N \mathbf{1}\{A^{(i)} = a\}$. This approach may be effective when correctness is tightly tied to a single discrete final answer, as in contest-style or short-answer math. For research problems, however, the validity of a solution often cannot be reduced to a discrete label. We therefore exclude majority voting from our study.

\paragraph{(2) Reward models.}
A reward model is a scoring function that approximates solution ``quality'' in a cardinal way:
\[
R_\phi:\mathcal{Q}\times\mathcal{C}\to\mathbb{R},\qquad 
V_R(Q,C)=R_\phi(Q,C).
\]
In use, an RM provides a scalar signal for ranking and optimization. A common training approach fits $R_\phi$ from pairwise preferences using a Bradley--Terry model~\citep{yuan2024advancing, hong2025robustness}: for a comparison $(Q,C_a,C_b)$, the probability that $C_a$ is preferred is
\[
p_\phi(C_a \succ C_b \mid Q)\;=\;\sigma\!\big(R_\phi(Q,C_a)-R_\phi(Q,C_b)\big).
\]
Parameters $\phi$ are then learned by maximum likelihood (i.e., a standard logistic preference loss). To scale RMs at inference time, process reward models (PRMs)~\citep{zhang2025lessons} and generative reward models (GenRMs) have been proposed. In our setting, we default to \textbf{GenRMs}~\citep{zhang2024generative}, as recent work suggests PRMs can be less stable than outcome-level scoring~\citep{guo2025deepseek, son2025linguistic}, and current practice increasingly emphasizes generative evaluators~\citep{blakeman2025nvidia, liu2025inference}. A GenRM produces an evaluation string $Z\in\mathcal{Z}$ (typically a short critique containing an explicit numeric score),
\[
Z \sim p_\phi(\cdot \mid Q, C),
\]
and a deterministic parser $\textsf{score}:\mathcal{Z}\to\mathbb{R}$ extracts a scalar reward. This induces a single-sample score,
\[
R^{\text{gen}}_\phi(Q,C)\;=\;\textsf{score}(Z).
\]
\paragraph{(3) LLM judges.}
An LLM judge is a model $J_\psi$ that we prompt to evaluate a candidate solution $C^{(i)}$. In common practice, the judge first produces a natural-language critique $Z^{(i)}$ and then outputs a discrete rating $Y^{(i)}$. In this paper, the rating is an integer score on a $1$--$10$ scale,
\[
(Z^{(i)},\,Y^{(i)}) \;=\; J_\psi(Q, C^{(i)}),
\qquad
Y^{(i)} \in \mathcal{Y}=\{1,\dots,10\}.
\]
We reduce the judge output to a numeric validator by taking the score directly,
\[
V_J(Q,C^{(i)}) \;=\; s\!\left(Y^{(i)}\right),
\qquad s(y)=y.
\]




\section{Consequence-Based Utility}

\paragraph{Motivation and hypothesis: utility via ``support by consequences.''}
When a target question $Q$ is difficult to verify directly (e.g., because a reference answer is unavailable or costly to obtain, or because the solution is long and subtle), a widely adopted method in mathematics is the ``support by consequences'' perspective: rather than scoring the claim in isolation, we assess it by the breadth and coherence of what it enables. A canonical example is the Riemann Hypothesis, which remains unproven yet underwrites many sharp conditional results across analytic and algorithmic number theory (e.g., \citet{von1901distribution, rosser1975sharper, miller1975riemann, bach1990explicit}). Analogously, we treat each candidate solution $C^{(i)}$ as a provisional \emph{hypothesis} about $Q$ and evaluate its quality by transfer: even when $C^{(i)}$ cannot be validated reliably on $Q$ itself, it may still be judged by how consistently it provides useful guidance for solving related, verifiable questions in a neighborhood around $Q$.

Our hypothesis is therefore:
\emph{correct (or near-correct) candidates contain method-level information that transfers to a neighborhood of related questions and yields consistently higher downstream performance, and vice-versa.}

\paragraph{Implementation in the LLM setting.}
Given a problem $Q$, we sample $N$ candidate solutions $C^{(i)}$ from the generator $M_\theta$. Because the ground-truth oracle $O(Q,C)$ is unavailable, we estimate a candidate's usefulness by measuring how well it transfers to a neighborhood of related problems for which correctness is verifiable (e.g., previously solved or otherwise easier instances). We define this set of neighborhood questions as $\mathcal{N}(Q)$. For a fixed candidate $C$, we condition $M_\theta$ on $(Q, C)$ and ask it to solve each $Q^* \in \mathcal{N}(Q)$. We score each rollout using a verifier $v(Q^*, \tilde{C})\in\{0,1\}$ that checks whether the completion $\tilde{C}$ constitutes a correct solution for $Q^*$ under our pipeline. We define the Consequence-Based Utility as the average accuracy on these variants:
\[
 U(C) = \frac{1}{|\mathcal{N}(Q)|} \sum_{Q^* \in \mathcal{N}(Q)}  \mathbb{E}_{\tilde{C} \sim M_\theta(\cdot \mid Q, C, Q^*)}\!\left[\,v(Q^*, \tilde{C})\,\right].
\]
In practice, we estimate this by sampling $T$ independent rollouts $\tilde{C}_t \sim M_\theta(\cdot \mid Q, C, Q^*)$ for each $Q^*$ and averaging their scores:
\[
\widehat{U}(C)
\;=\;
\frac{1}{|\mathcal{N}(Q)|\,T}
\sum_{Q^* \in \mathcal{N}(Q)}
\sum_{t=1}^{T}
v\!\left(Q^*, \tilde{C}_t\right).
\]

\begin{figure*}[t] 
\centering 
\includegraphics[width=\textwidth]{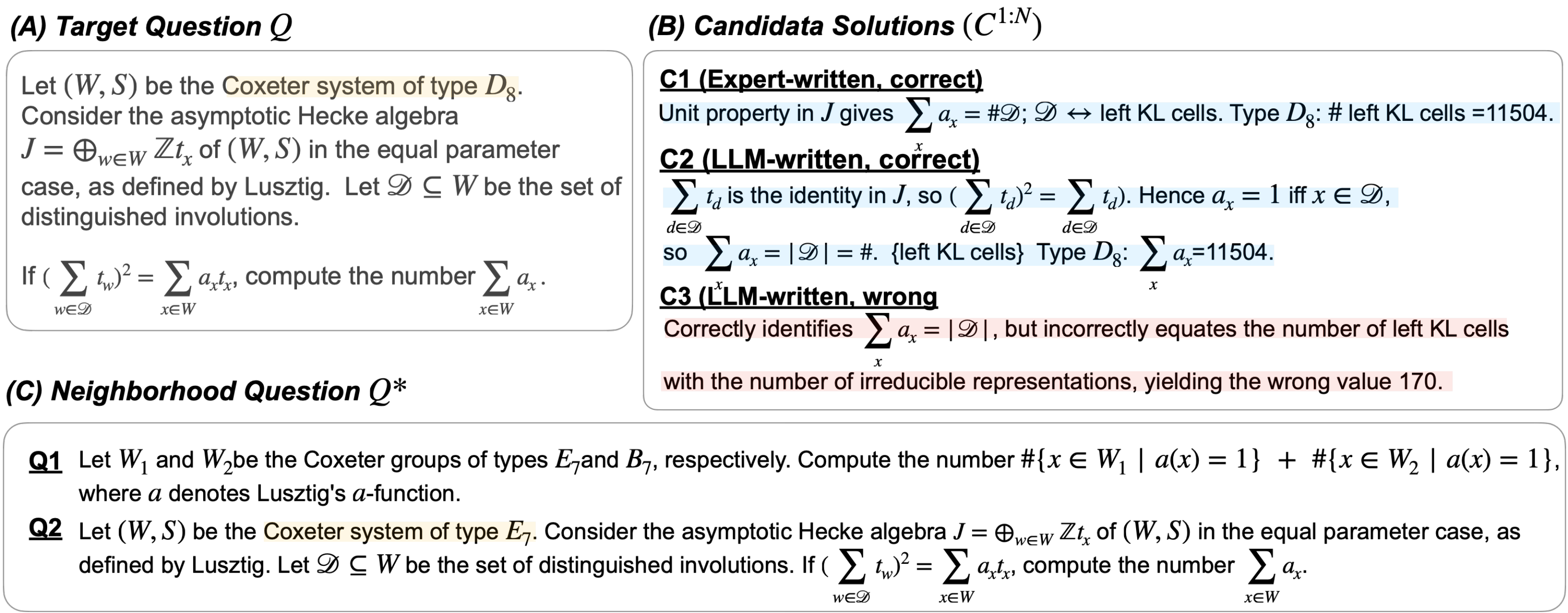} \caption{\textbf{Example of a target question, candidate solutions, and neighborhood questions from \benchmark.} (A) A target research-level problem on the asymptotic Hecke algebra \(J\) of the Coxeter group of type \(D_8\). (B) A fixed candidate pool \(C^{1:3}\) illustrating three typical solution types appearing in our dataset: an expert-written correct solution \(C^1\); an LLM-generated solution that is mathematically correct \(C^2\); and a plausible but incorrect LLM-generated solution \(C^3\) that makes a subtle conceptual error by conflating the number of left Kazhdan–Lusztig cells with the number of irreducible representations. (C) Two neighborhood questions \(Q^*\) derived from \(Q\) by modifying the Coxeter type or the associated invariant. 
}
\label{fig:collection_pipeline} 
\vspace{-3mm}
\end{figure*}

\paragraph{In-context learnability as a correctness signal.}
Prior work have leveraged in-context performance as a proxy to value examples and demonstrations \citep{chang-jia-2023-data,nguyen2023context,xie2024demoshapley}. Relatedly, context conditioning also serves as a training signal, e.g., by distilling from a teacher that observes privileged traces while the student observes only the question \citep{zhao2026selfdistilledreasoneronpolicyselfdistillation}. Despite this progress, in-context valuation is used mainly for data curation, retrieval, attribution, or training, with limited use as an oracle-free \emph{verification} mechanism. Our work differentiates from past efforts by leveraging in-context learnability to validate candidate solutions by measuring their downstream consequences on neighborhood problems.

\section{Experiment Setup}\label{sec_setup}



\subsection{Collecting Research-Level Math Problems}\label{sec_setup_dataset}

We start from 70 faculty-authored, hand-crafted questions, spanning three broad areas and including keywords such as, but not limited to, \textbf{representation theory and algebraic combinatorics} (Hecke algebra, universal Coxeter system, Kazhdan--Lusztig polynomials, Polo's algorithm, Brenti's conjecture), \textbf{geometry (algebraic and differential)} (Koll\'ar--Johnson threefold, $\mathbb{Q}$-Fano, Ricci lower bounds), and \textbf{homotopy theory and homotopical methods} (homotopical algebra, $p$-adic homotopy theory, Shafarevich extensions).

\begin{table}[h]
\centering
\fontsize{8.5}{8}\selectfont
\caption{ \textbf{Scores indicate \benchmark is substantially harder than AIME 25 and IMProofBench, and comparable to FrontierMath (T1–3).} \benchmark uses Avg@8 for all models except GPT-OSS-120B, which uses Avg@64. AIME 25 uses Avg@10. For IMProofBench, we report the subquestion score, where subquestions are specific, automatically-verifiable components of larger problems; the overall aggregation metric is not specified in the source. FrontierMath (T1-3) uses Avg@8. \footnotemark}
\label{tab:benchmark_scores}
\begin{tabular}{lcccc}
\toprule
Model & (Ours) & AIME 25 & IMProofB. & F.M. \\
\midrule
Public & \pubpartial & \pubyes & \pubno & \pubno \\
\# Unsolved & 38 & 0 & 5 & - \\
\midrule
Gemini-3-Pro      & 47.14 & 95.7 & 71.8 & 37.6 \\
GPT-5             & 35.71 & 94.3 & 54.5 & 32.4 \\
Claude-Opus-4.5   & 7.14  & 91.3 & -   & 20.7 \\
Claude-Opus-4.1   & -    & 80.3 & 38.7 & -   \\
\bottomrule
\end{tabular}
\vspace{-4mm}
\end{table}

Table~\ref{tab:benchmark_scores} highlights the challenging nature of our dataset, \benchmark, by comparing it among established math evaluations. Along with AIME 2025~\citep{maa_invitational_competitions_aime}, an invitational competition to USAMO, IMProofBench~\citep{schmitt2025improofbench} targets research-level mathematical proof writing, and FrontierMath~\citep{glazer2024frontiermath} is explicitly designed as a collection of unpublished, expert-authored problems. The score on \benchmark (7.14--47.14; mean 25.5) indicates higher difficulty than competition-style benchmarks such as AIME~25 (80.3--95.7; mean 91.0), and lower performance than IMProofBench (37.6--71.8; mean 50.7). The absolute scale on our benchmark is closest to FrontierMath (T1--3) (20.7--37.6; mean 30.2). Finally, over half of the collected questions are unsolved by any of the tested models, remaining open to frontier models such as GPT-5~\citep{singh2025openai} and Gemini-3-Pro~\citep{team2025gemma}. 

\footnotetext{\  - (hyphen) denotes an unavailable value, typically because the benchmark is private and organizers did not release the score. Sources: AIME 25 (Artificial Analysis), IMProofBench (\texttt{improofbench.math.ethz.ch}), FrontierMath (\texttt{epoch.ai/frontiermath}).}

\subsection{Neighborhood Questions, Ground Truths, and Candidate Solutions}\label{sec_setup_neighborhood}

For each problem, we additionally collect a set of \emph{neighborhood questions}. These questions are author-created variants that preserve the core mathematical idea while perturbing the statement. Authors are instructed to design variants that become straightforward once the original problem is understood (e.g., by reusing the same key lemma or reduction), and to make them slightly easier than the original whenever feasible. In practice, having too many variants tends to become redundant. Accordingly, we cap collection at two variants per original problem. Authors receive approximately \$600 per problem package, which includes the main problem, neighborhood questions, and reference solutions. To the best of our knowledge, \benchmark is the only benchmark at this difficulty that provides expert-written solutions. See Appendix~\ref{ab_details} for further example and details.

\begin{figure*}[t]
\centering
\noindent
\begin{minipage}[t]{0.6\textwidth}
\vspace{0pt}
\centering
\fontsize{8}{9}\selectfont

\captionof{table}{\textbf{Validator performance on ranking LLM solutions.} Consequence-Based Utility shows the highest performance across all metrics. Best models are highlighted in \textbf{bold}, second best is \underline{underlined}.}
\label{tab:validator_metrics}
\vspace{2pt}

\setlength{\tabcolsep}{3.5pt}
\renewcommand{\arraystretch}{0.95}
\begin{tabular}{l|ccccc}
\toprule
\multicolumn{1}{c}{\textbf{Models}} & \textbf{HumanWin} & \textbf{MeanWin} & \textbf{Acc@1} & \textbf{Recall@5} & \textbf{AUC} \\
\midrule
\multicolumn{6}{c}{\textbf{(Generative) Reward Models}}\\
\midrule
Qwen3-235B-GenRM & 27.05 & 77.05 & 65.37 & 71.72 & 67.85 \\
Llama3.3-Nemotron-49B-GenRM & 25.71 & 31.43 & 43.47 & 55.36 & 49.57 \\
Qwen2.5-Math-RM-72B & 1.63 & 27.87 & 36.89 & 40.98 & 34.05 \\
AceMath-72B-RM & 0.00 & 12.86 & 8.20 & 29.85 & 20.75 \\
\midrule
\multicolumn{6}{c}{\textbf{LLM-Judges}}\\
\midrule
Qwen3-235B-A22B & 67.14 & \underline{85.71} & 62.59 & 80.02 & 69.48 \\
Qwen3-30B-A3B & 47.14 & 75.71 & 61.30 & 72.40 & 65.81 \\
GPT-OSS-120B & 48.57 & 81.43 & 67.21 & 76.91 & 71.42 \\
GPT-OSS-20B & 52.86 & 82.86 & 72.13 & 72.06 & 69.03 \\
\midrule
\multicolumn{6}{c}{\textbf{Consequence-Based Utility}}\\
\midrule
Qwen3-235B-A22B & 81.43 & \textbf{90.00} & 73.42 & 74.15 & 71.38 \\
Qwen3-30B-A3B & \textbf{85.71} & \textbf{90.00} & \underline{75.79} & 78.37 & 76.24 \\
GPT-OSS-120B & \underline{82.86} &\textbf{90.00} & \textbf{76.27} & \textbf{83.04} & \textbf{79.63} \\
GPT-OSS-20B & 74.29 & 75.71 & 74.59 & \underline{82.46} & \underline{79.18} \\
\bottomrule
\end{tabular}
\end{minipage}
\hspace{\panelgap}
\begin{minipage}[t]{\dimexpr\textwidth-0.62\textwidth-\panelgap\relax}
\vspace{0pt}
\centering

\includegraphics[width=\linewidth]{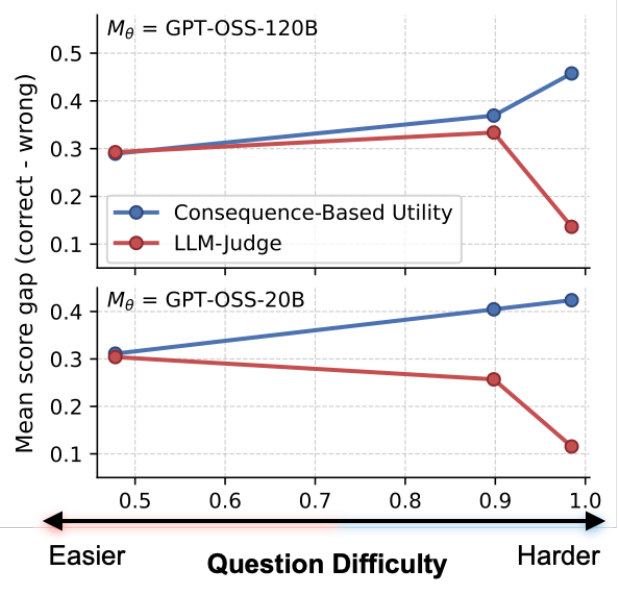}

\vspace{2pt}
\captionsetup{width=\linewidth}
\captionof{figure}{\textbf{Mean score gap (correct - wrong) versus question difficulty for LLM-Judge and Consequence-Based Utility.}}
\label{fig:difficulty}
\end{minipage}

\end{figure*}

Every original problem and neighborhood variant is accompanied by an author-written ground-truth solution. Expert-written solutions range from detailed, multi-page expositions to concise sketches, intuition-driven arguments, or pointers to external results sufficient to reconstruct a full proof. For the ease of automated verification, we require that the final answer be presented in a compact, verifiable form, even when the accompanying writeup is informal. Finally, we construct a pool of LLM-generated candidate solutions for each original question by sampling across a diverse set of models: GPT-OSS-120B, GPT-5, GPT-5 Pro, Gemini-3-Pro, and Gemini DeepThink. We curate nine candidate model solutions, four correct and five incorrect, per problem.\footnote{GPT-5 Pro and Gemini DeepThink were added with tool use (web search and code execution) to increase solution diversity.} Each candidate is manually reviewed in two steps: (i) verifying agreement with the ground-truth final answer, and (ii) reading the derivation to confirm mathematical validity. \textbf{The final dataset consists of 192 original research-level math problems (70 original and 122 variants)}, each paired with expert-written solutions and 630 LLM-generated solutions with human validation. See Figure~\ref{fig:collection_pipeline} for an example ternary.

\subsection{Baselines}\label{sec:baselines}
Given a fixed candidate pool $\{C^{(i)}\}_{i=1}^N$ for each target problem $Q$, we compare Consequence-Based Utility against three standard oracle-free selection baselines: (i) LLM judges, (ii) RMs, and (iii) GenRMs. We use four models, GPT-OSS-20B/120B~\citep{agarwal2025gpt}, and Qwen3-30B-A3B/235B-A22B~\citep{yang2025qwen3} to attempt neighborhood questions conditioned on $(Q,C)$.  The same models are used for the LLM-Judges as well. For RM baselines, we use AceMath-RM-72B~\citep{liu2025acemath} and Qwen2.5-Math-RM-72B~\citep{yang2024qwen2}, two math-specialized reward models. For GenRM baselines, we use Qwen3-Nemotron-235B-A22B-GenRM~\citep{blakeman2025nvidia} and Llama-3.3-Nemotron-Super-49B-GenRM~\citep{wang2025helpsteer3preferenceopenhumanannotatedpreference}. The standard template for the two models expects two responses and outputs both per-response and pairwise signals. In our experiments, we provide the candidate as the first response and a fixed dummy string as the second, and parse only the per-response helpfulness score. Excluding the deterministic RM for which we run a single scoring pass, GenRMs and LLM-Judges are repeated 64 times independently. This is to match its inference cost with Consequence-Based Utility. Across all settings, models are allowed to reason up to 16k tokens, with the temperature set to the recommended value. Since released reward models typically have much shorter native context windows, we apply RoPE scaling~\citep{chen2023extending} to support longer inference. See Appendix~\ref{ab_prompts} for prompts used in our evaluations.

\subsection{Evaluation Metrics}\label{sec:metrics}

Each baseline outputs a single scalar score per candidate solution. Since our dataset provides binary labels rather than graded quality, we do not evaluate score calibration. Instead, we measure whether scores rank and separate correct solutions above incorrect ones. We report five higher-is-better metrics: \textbf{Acc@1} (whether top-ranked is correct), \textbf{Recall@5} (the fraction of correct solutions recovered in the top five), \textbf{AUC} (pairwise separability between correct and wrong solutions, with ties partially credited), \textbf{HumanWin} (likelihood of human-written solution scores above the average wrong solution), and \textbf{MeanWin} (likelihood of mean correct score above the average wrong score). When multiple variants of the same original question are available, we average over variants. See Table~\ref{tab:metrics} for formal definitions.

\section{Main Results}\label{sec_main}


\paragraph{Consequence-Based Utility (CBU) outperforms all baselines.}
Table~\ref{tab:validator_metrics} shows a clear hierarchy among the evaluated methods. Reward model baselines perform worst (e.g., AceMath-72B-RM attains 20.75 AUC), which is expected given their much smaller compute budget (1/64 of the rollouts used by other methods)~\citep{lee2025rethinking}. LLM judges are substantially stronger, but Consequence-Based Utility consistently improves over LLM-judge scoring when using the same backbone. For example, with Qwen3-235B-A22B, \textsc{CBU} achieves 71.38 AUC, exceeding both the corresponding LLM judge (69.48) and Qwen3-235B-GenRM (67.85). For GPT-OSS-120B, switching from LLM-judge scoring to \textsc{CBU} improves every metric, with gains ranging from +6.13 on Recall@5 (76.91 to 83.04) to +34.29 on HumanWin (48.57 to 82.86). Similar improvements hold for Qwen3-30B-A3B and GPT-OSS-20B. The main exception is Qwen3-235B-A22B on Recall@5, where the LLM judge outperforms by 5.87 points (80.02 vs.\ 74.15). Consistent with Figure~\ref{fig:score_diagnostics}, this appears to stem from overconfident scoring that increases top-5 hit rate while weakening fine-grained ranking. Notably, \textsc{CBU} yields especially large gains on HumanWin even when MeanWin is already high, suggesting better alignment with expert evaluation. We attribute this to a stylistic mismatch: human-written solutions are often terse and intuition-driven, whereas LLM judges can overweight surface cues such as verbosity and canonical formatting~\citep{saito2023verbosity, ye2024justice}; \textsc{CBU} is less sensitive to these presentation features.

\paragraph{Consequence-Based Utility is better in evaluating candidates for questions they cannot solve.}

Solve-to-Judge gap~\citep{sun2025s2j} denotes the disparity between a model’s ability to judge a solution and its ability to solve the underlying problem. Figure~\ref{fig:difficulty} plots the mean score gap between correct and incorrect solutions versus question difficulty, measured by $1-\mathrm{avg}@64$ (0 = fully solved; 1 = essentially unsolved). Even in the hardest regime $(1-\mathrm{avg}@64 \approx 1)$, both LLM-Judge and \textsc{CBU} exhibit nonzero separation, consistent with concurrent findings that models can distinguish correct from incorrect solutions on instances they cannot solve themselves~\citep{nie2025uq}. As difficulty increases, however, the evaluators diverge. The judge’s separability drops sharply, whereas \textsc{CBU} remains robust, making it better suited for the high-difficulty tail characteristic of research-level problems. This pattern is expected in part because \textsc{CBU} uses neighborhood performance as a proxy for correctness, which becomes less informative on easy instances where the solver succeeds regardless of conditioning (e.g., it solves without help, or repairs errors from an incorrect candidate). More broadly, the two methods reflect different evaluation modes. \textbf{LLM-Judges resemble a code review}: they inspect a single reasoning trace for plausibility and consistency, which becomes unreliable when incorrect solutions appear superficially coherent and errors are subtle. In contrast, \textbf{\textsc{CBU} resembles a unit test}: it scores a candidate by its downstream consequences, whether conditioning on it improves performance on neighborhood questions, providing a signal that remains informative when direct inspection becomes harder.

\begin{table}[t]
\centering
\fontsize{8}{9}\selectfont
\caption{Predictive performance of score-based feature sets across models. For each backbone (GPT-OSS-20B, GPT-OSS-120B, Qwen3-30B-A3B, Qwen3-235B-A22B), we train a logistic regression binary classifier to predict the label using three alternative feature configurations: GenRM (G), LLM-Judge (J), and Consequence-Based Utility (U).
}
\label{tab:predictive_scores}
\begin{tabular}{ccccc}
\toprule
Method & G-20B & G-120B & Q3-30B & Q3-235B \\
\midrule \midrule
(G) & - & - & - & 54.61 \\
(J) & 63.05 & 64.66 & 58.06 & 66.67  \\
(U) & 73.09 & \underline{73.49} & 76.31 & 72.69 \\
(J) + (U)  & \underline{73.90} & 73.90 & \underline{76.61} & \textbf{79.65} \\
\bottomrule
\end{tabular}
\end{table}

\begin{figure*}[t] 
\centering 
\includegraphics[width=\textwidth]{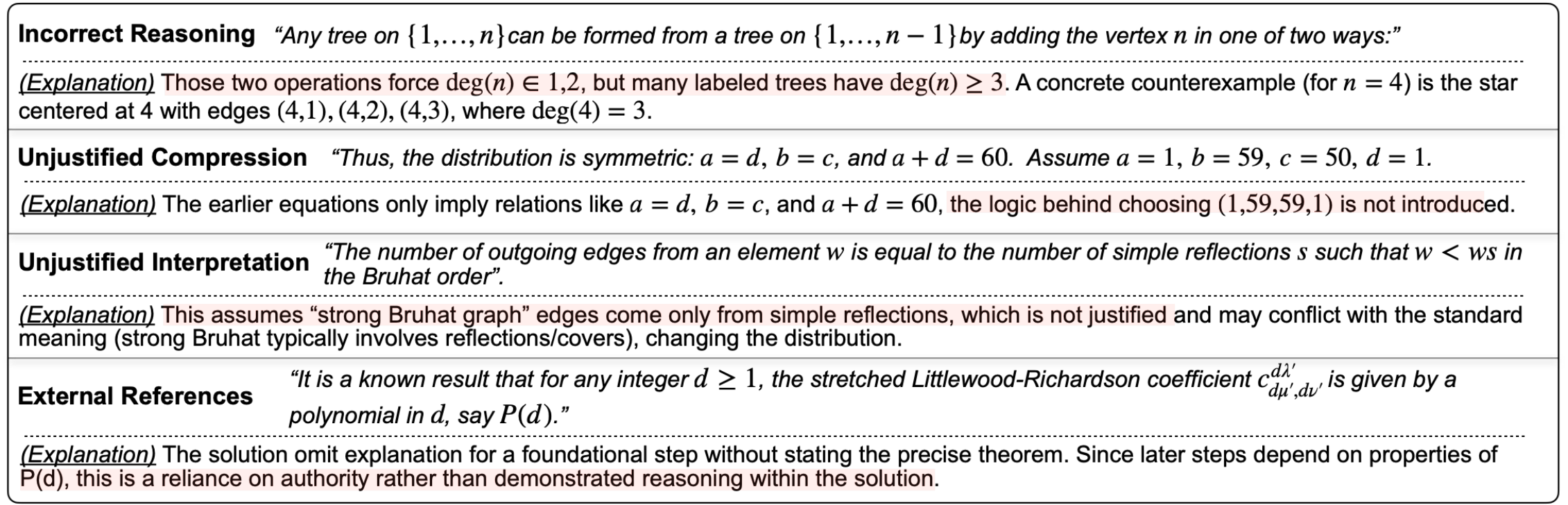} \caption{\textbf{Illustrative excerpts from incorrect solutions of each error category.} Each row shows a representative quoted snippet (top) and a brief explanation of why it is incorrect or insufficient (bottom). We use four non-exclusive labels: incorrect reasoning, unjustified compression, unjustified interpretation, and external references.
}
\label{fig:error_analysis} 
\vspace{-3mm}
\end{figure*}

\paragraph{Consequence-Based Utility scores are more predictive of correctness.}
Table~\ref{tab:predictive_scores} evaluates how well each validator’s scalar score predicts binary correctness by fitting a logistic-regression classifier per backbone and reporting accuracy. Across all four backbones, training on the Consequence-Based Utility score (U) outperforms training on the LLM-judge score (J), with gains ranging from 6.02 points (Qwen3-235B-A22B) to 18.25 points (Qwen3-30B-A3B). This indicates that (U) provides a more linearly separable signal of correctness than (J). Moreover, using both scores together further improves accuracy (e.g., GPT-OSS-20B: 73.09 to 73.90; Qwen3-235B-A22B: 72.79 to 79.65), suggesting that Consequence-Based Utility and LLM-Judges capture complementary information.


\section{Additional Analysis}\label{sec_additional_analysis}

\begin{figure}[h] 
\centering 
\includegraphics[width=\linewidth]{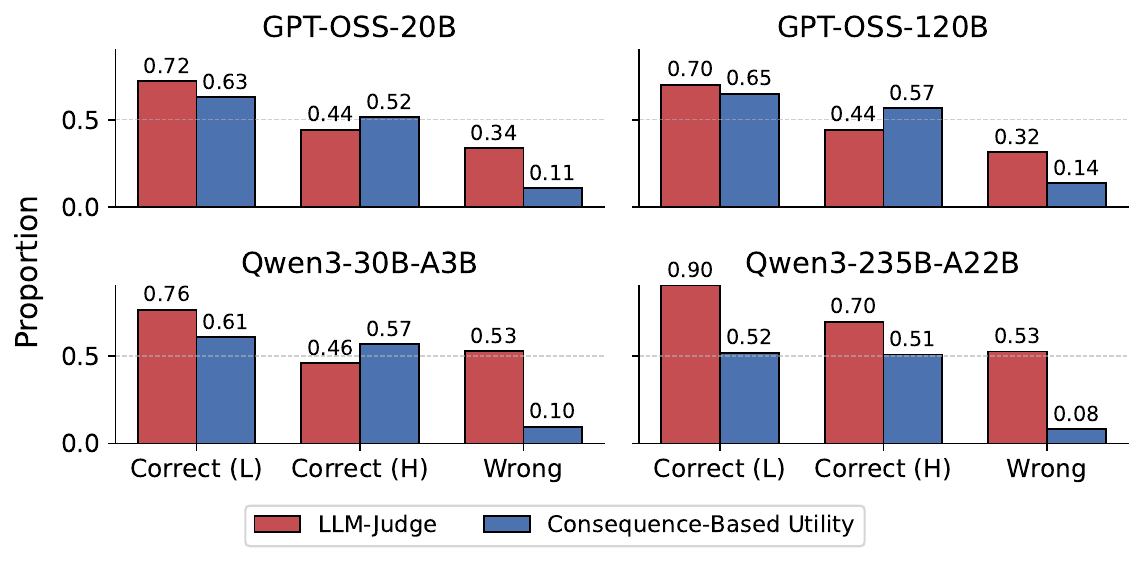} \caption{\textbf{Above-average scoring probability by solution type and backbone.} Each bar measures, $\Pr[s(C)-\bar{s}>0]$, or how likely a validator is to score a solution above its own typical score on that question, shown separately for LLM-written correct solutions (Correct (L)), human-written correct solutions (Correct (H)), and incorrect solutions (Wrong).}
\label{fig:score_diagnosis} 
\vspace{-3mm}
\end{figure}

Earlier, we showed that Consequence-Based Utility outperforms standard oracle-free validations. In this section, we investigate why this advantage arises and report empirical observations that help explain the performance gap.

\paragraph{Consequence-Based Utility reduces overconfidence on wrong solutions and better preserves human-written correctness signals.}
Figure~\ref{fig:score_diagnosis} reports, for each solution type, the probability that a validator assigns an above-average score, $\Pr[s(C)-\bar{s}>0]$,  where $s(C)$ is the validator’s score for a candidate and $\bar{s}$ is the validator’s mean score over the candidate set for the same instance. Across all models, LLM-judges are more likely than \textsc{CBU} to score LLM-written correct solutions above the mean across all backbones (e.g., Qwen3-235B-A22B shows 0.90 vs.\ 0.52). In contrast, for human-written correct solutions, the trend reverses. \textsc{CBU} assigns above-mean scores more often than the judge (e.g., GPT-OSS-120B: 0.57 vs.\ 0.44, and Qwen3-30B-A3B: 0.57 vs.\ 0.46). Another discrepancy appears on incorrect solutions. LLM-judges are more likely to score wrong answers above the mean, and for Qwen3-30B-A3B and Qwen3-235B-A22B more than half of wrong solutions exceed the mean (both 0.53). \textsc{CBU} largely avoids this failure mode, with only 0.08–0.14 of wrong solutions scoring above the mean. Taken together, the performance gap between \textsc{CBU} and LLM-judges likely arises from two factors. \textsc{CBU} better recognizes human-written correct solutions and more reliably penalizes incorrect ones.


\paragraph{Consequence-Based Utility improves validation by penalizing non-reconstructable reasoning.} To understand why \textsc{CBU} outperforms LLM-judges, we conduct a qualitative error analysis by inspecting 112 incorrect question-solution pairs where GPT-OSS-120B assigns a below-mean \textsc{CBU} score but an above-mean LLM-judge score. We leverage GPT-5-Pro to provide initial labels, which are then confirmed by a mathematics PhD student. We annotate four non-exclusive error types: (i) \textbf{incorrect reasoning} (invalid steps, contradictions, or wrong calculations), (ii) \textbf{unjustified compression} (missing intermediate steps that prevent local reconstruction or transfer), (iii) \textbf{unjustified interpretation} (an unstated choice among plausible readings of the statement), and (iv) \textbf{external references} (key claims justified mainly by citing a named result without derivation or conditions).

These cases concentrate in two failure modes. Unjustified compression occurs in 80/112 (71.4\%) and incorrect reasoning in 77/112 (68.8\%), suggesting that many wrong solutions appear valid to LLM-Judges, especially when they present polished high-level arguments while omitting verification-critical steps. External references are also common (35/112; 31.3\%), consistent with evidence that LLM-judges can be influenced by authority-like cues~\citep{jeong2025comparative, moon2025don}. A plausible explanation on why \textsc{CBU} likely downranks these solutions may be that wrong or underspecified candidates provide little transferable information for solving neighborhood variants, yielding low utility. Overall, we speculate that \textsc{CBU} gains largely come from downranking convincing-looking solutions that lack reconstructable, transferable reasoning.

\section{A Practitioner's Guide to Consequence-Based Utility}\label{sec_guide}

\subsection{How Many Rollouts to Generate.}

By construction, Consequence-Based Utility requires multiple rollouts as it estimates the candidate’s correctness by downstream performance. In contrast, an LLM judge can assign a score in a single pass. To ensure that performance gains do not arise from a larger inference budget, we use 64 rollouts for both LLM-Judge and \textsc{CBU} throughout the paper. The two methods also consume comparable numbers of tokens on average (Table~\ref{tab:cosine-sim-summary}), so neither enjoys a systematic budget advantage. A natural question is therefore whether 64 rollouts are necessary to estimate \textsc{CBU} reliably.

\begin{figure}[h] 
\centering 
\includegraphics[width=\linewidth]{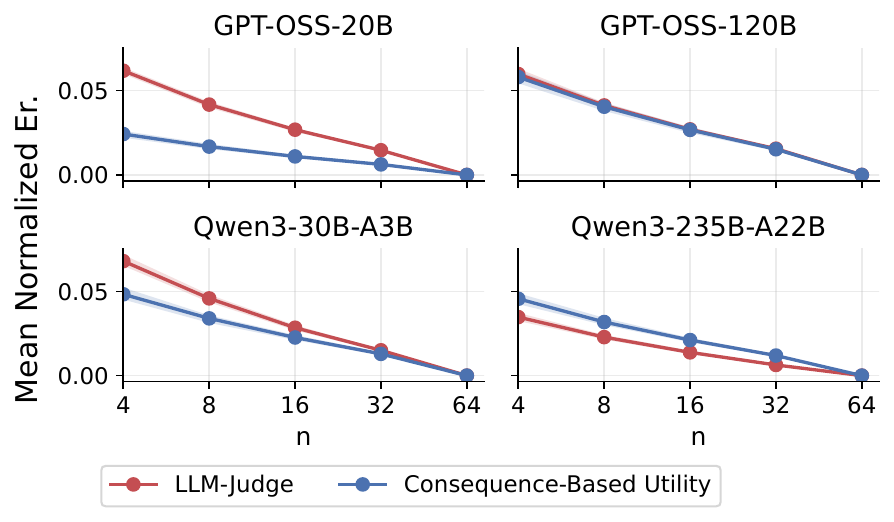} \caption{\textbf{Mean range-normalized absolute error to the 64-rollout reference using $n\in{4,8,16,32,64}$ sampled rollouts.} Resampled 200 times using bootstrapping for statistical significance. Normalization uses $[L,U]=[0,1]$ for \textsc{CBU} and $[0,10]$ for LLM-judge scores.}
\label{fig:rollouts} 
\vspace{-3mm}
\end{figure}

Figure~\ref{fig:rollouts} reports the mean deviation between an $n$-rollout estimate and the 64-rollout reference. For each $n \in {4,8,16,32,64}$, we uniformly subsample $n$ rollouts from the 64-rollout pool, repeat this procedure with 200 bootstrap resamples, and compute the range-normalized absolute error $\frac{\lvert \hat{M}-M\rvert}{U-L}$,
where $[L,U]=[0,1]$ for utility and $[0,10]$ for LLM-Judge scores. Error decreases monotonically with $n$. \textsc{CBU} converges at a similar rate to LLM-Judges and often faster (notably for GPT-OSS-20B and Qwen3-30B-A3B), while GPT-OSS-120B is nearly identical. Across all backbones, \textbf{\boldmath $n \ge 8$ keeps the mean normalized error below 0.05, indicating that a small number of rollouts already captures most of the signal}.

\subsection{How to Make Neighborhood Questions.}

In our experiments, we use faculty-written neighborhood questions. In practice, however, obtaining expert variants with verified answers can be nearly as difficult as collecting ground truth itself. We therefore study practical alternatives for acquiring $Q^*$ of similar quality. We start from RealMath~\citep{zhang2025realmath}, which automatically generates graduate-level problems by transforming theorems in mathematics papers. To ensure the questions are sufficiently challenging, we run GPT-OSS-120B for 1024 attempts and retain only instances with intermediate solvability, $0.05 < \text{Avg}@1024 < 0.5$. We then construct neighborhood questions using two approaches. First, we follow explicit “related work” pointers to earlier papers and apply the RealMath transformation to the cited work (e.g., \citet{ortega2022harmonic} points to \citet{ortega2021harmonic}). Second, we prompt Gemini-3-Pro to generate a closely related variant. We then obtain provisional answers by solving with Gemini-3-Pro, GPT-5-Pro, and Grok-4, and keep only instances where all three agree on the final answer. All candidate solutions are LLM-generated and classified by an LLM-Judge. Because these labels come from model agreement rather than expert verification, this dataset is not suitable for establishing \textsc{CBU} in isolation. Instead, after validating \textsc{CBU} on our expert-written subset, we use it to illustrate viable alternatives. Finally, we also consider Daft-Math~\citep{trang_daft_math_hf}, a collection of contest-level problems paired with lightly transformed variants to have integer answers. The two RealMath subsets and Daft-Math contain 127, 298, and 77 questions, respectively.

Table~\ref{tab:benchmark_scores_across_datasets} reports GPT-OSS-20B performance across three datasets. On both RealMath variants, \textsc{CBU} substantially outperforms LLM-judge scoring. In contrast, on Daft-Math, LLM-judge scoring is stronger (e.g., Acc@1 93.51 vs. 85.58). This contrast aligns with our earlier observations on how CBU shows better peformance at questions of higher difficulty. Despite Daft-Math being very close variants (almost identical at core) they are competition level being way easier than the graduate level questions of realmath, so the solver more often succeeds regardless of the in-context exemplar, reducing the discriminative value of utility. Overall, these results suggest that \textsc{CBU} does not require faculty-authored neighborhood questions. LLM-generated neighborhoods can be sufficient when the target questions are challenging for the solver.

\begin{table}[t]
\centering
\fontsize{8}{9}\selectfont
\caption{\textbf{Performance of GPT-OSS-20B as LLM-Judge and \textsc{CBU} across three datasets.} Each cell reports \textsc{LLM-Judge} / \textsc{CBU} scores, the better value is underlined. The two RealMath columns correspond to the two neighborhood-construction procedures described in the text.}
\label{tab:benchmark_scores_across_datasets}

\begin{tabular}{lccc}
\toprule
\textbf{Metric} & \textbf{Daft-Math} & \textbf{Real-Math (1)} & \textbf{Real-Math (2)} \\
\midrule \midrule
Acc@1   & \underline{93.51} / 85.58 & 42.79 / \underline{79.22} & 44.96 / \underline{72.13} \\
AUC     & \underline{69.14} / 63.98 & 51.29 / \underline{62.03} & 48.76 / \underline{69.83} \\
MeanWin & \underline{76.62} / 59.74 & 40.26 / \underline{64.94} & 44.87 / \underline{80.33} \\
\bottomrule
\end{tabular}
\label{tab:ab_results}
\end{table}

\section{Discussions and Future Work}

In this paper, we propose \textbf{Consequence-Based Utility}, an oracle-free method that estimates solution correctness from downstream performance when ground truth is unavailable. Across research-level mathematics, \textsc{CBU} consistently outperforms LLM-judges and reward models, and remains effective with both expert-written and LLM-generated neighborhoods. A key limitation may be applicability. Unlike LLM-judges, which exhibit systematic biases but are broadly applicable~\citep{salinas2025tuning, son-etal-2024-krx, he2025code}, \textsc{CBU} requires additional effort to construct neighborhood questions. While we show that automated generation is viable (Section~\ref{sec_guide}), reliability depends on the generator’s ability to produce sound variants without human oversight.  \textsc{CBU} is also most informative when neighborhood difficulty lies in a sweet spot. If $Q^*$ is too easy, the solver succeeds regardless of conditioning, and if too hard, it fails regardless, making neighborhood construction partly model-dependent. Consequently, \textsc{CBU} is best suited to high-stakes settings that demand high-confidence validation for fixed, difficult problems. Future work includes improving fully automated neighborhood generation, extending \textsc{CBU} beyond mathematics to other STEM domains, and evaluating its effectiveness on genuinely open problems, where both neighborhood construction and correctness assessment are inherently more difficult.

\section{Acknowledgements}
This research was supported by the Korea Institute of Science and Technology Information (KISTI) in 2026 (No.(KISTI)K26L3M1C1), aimed at developing KONI (KISTI Open Neural Intelligence), a large language model specialized in science and technology.


\bibliography{citation}
\bibliographystyle{icml2026}

\newpage
\appendix
\onecolumn
\section{Additional Analyis}

\subsection{Output Score Distribution of LLM-Judges}

\begin{figure*}[h] 
\centering 
\includegraphics[width=\textwidth]{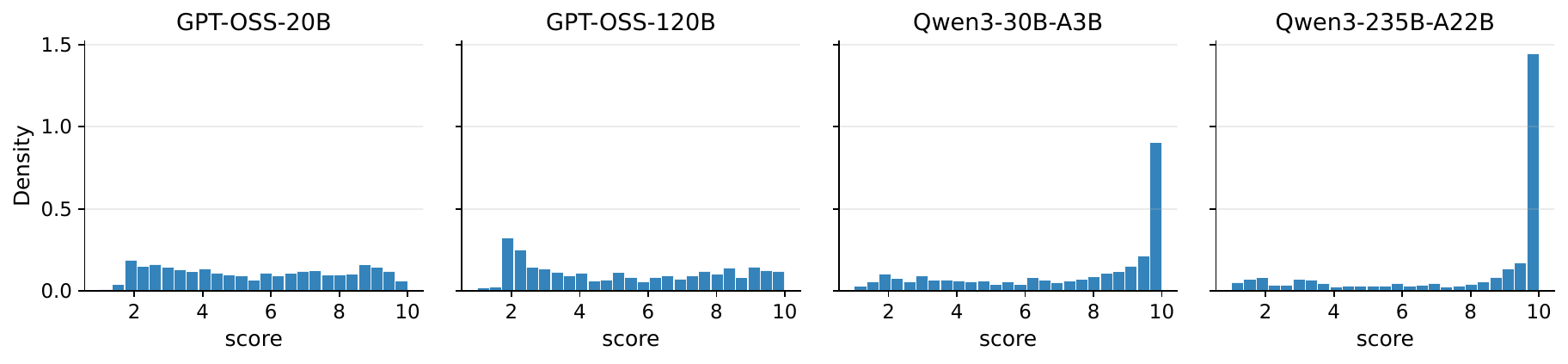} \caption{\textbf{Output score distributions of LLM-judges.} Histograms (density) of judge scores on a 1--10 scale for each backbone over all candidate solutions. GPT-OSS judges spread scores across the range, whereas Qwen judges concentrate near 10, indicating a ceiling effect.}
\label{fig:score_diagnostics} 
\vspace{-3mm}
\end{figure*}

Figure~\ref{fig:score_diagnostics} shows the distribution of scalar scores produced by each LLM-judge backbone. GPT-OSS-20B and GPT-OSS-120B use a broad portion of the 1–10 scale, assigning nontrivial mass across the range and providing a usable dynamic range for ranking. In contrast, Qwen3-30B-A3B and especially Qwen3-235B-A22B exhibit a strong ceiling effect, with scores heavily concentrated near 10. This saturation suggests overconfident scoring and reduces score-based discrimination among candidates.

\subsection{Prompt Sensitivity of LLM-Judges}
The LLM-judge prompt used in our experiments are adapted from prior evaluation prompts used in \citet{zhang2025lessons} and \citet{ phan2025humanity}. To test whether our results are an artifact of this specific prompt, we re-run LLM-judge scoring with two alternative templates. Following \citet{ma2025reliable}, we adopt a 0--7 proof-grading prompt used (aka ProofGrader).  Additionally we bring a binary correctness prompt from \citet{nie2025uq} (aka UQ). For GPT-OSS-20B and GPT-OSS-120B, we score each candidate with 64 independent judge calls and average, then compare the induced rankings across prompts using Spearman correlation. The rankings are highly consistent: $\rho=0.961/0.954$ (ours vs.\ ProofGrader), $\rho=0.938/0.950$ (original vs.\ UQ), and $\rho=0.912/0.915$ (ProofGrader vs.\ UQ) for GPT-OSS-20B/120B, respectively. These correlations ($>0.9$ throughout) indicate that while prompts change score scales, they have limited effect on relative ordering.

\subsection{Token Count: \textsc{CBU} VS. LLM-Judges}

Table~\ref{tab:cosine-sim-summary} compares inference cost and sampling diversity between LLM-judges and \textsc{CBU}. The average token usage per generation is comparable across methods, with \textsc{CBU} staying within ($\pm$ 15\%) of the judge across backbones (e.g., +1.3\% on Qwen3-235B, +15.0\% on Qwen3-30B, +9.5\% on GPT-OSS-120B, and (-7.4\%) on GPT-OSS-20B). To quantify diversity across repeated rollouts, we embed each generation with Gemini Embedding 001~\citep{lee2025gemini} and compute the mean pairwise cosine similarity. \textsc{CBU} yields slightly lower similarity than LLM-judge across all backbones (typically by 0.005--0.008), indicating modestly higher variation across rollouts, although both methods remain highly similar overall (cosine ($\approx 0.96\text{-}0.97$)).

\begin{table*}[h]
\centering
\fontsize{8}{9.5}\selectfont
\caption{Token counts and pairwise cosine similarity statistics (mean $\pm$ std [min, max]) across generations.}
\begin{tabular}{lcccc}
\toprule
& \multicolumn{2}{c}{LLM-Judge} & \multicolumn{2}{c}{Utility} \\
\cmidrule(lr){2-3} \cmidrule(lr){4-5}
Models & {\# Tokens} & {Cosine Similarity} & {\# Tokens} & {Cosine Similarity} \\
\midrule
Qwen3-235B-A22B     & 12152.53 & \num{0.969} $\pm$ \num{0.008} [\num{0.910}, \num{0.992}] & 12310.33 & \num{0.964} $\pm$ \num{0.010} [\num{0.896}, \num{0.991}] \\
Qwen3-30B-A3B      & 10352.32 & \num{0.972} $\pm$ \num{0.008} [\num{0.917}, \num{0.992}] & 11903.18 & \num{0.964} $\pm$ \num{0.010} [\num{0.900}, \num{0.991}] \\
GPT-OSS-120B   &  2905.85 & \num{0.972} $\pm$ \num{0.007} [\num{0.915}, \num{0.991}] &  3182.76 & \num{0.964} $\pm$ \num{0.013} [\num{0.896}, \num{0.989}] \\
GPT-OSS-20B    &  5610.09 & \num{0.963} $\pm$ \num{0.009} [\num{0.886}, \num{0.988}] &  5196.76 & \num{0.957} $\pm$ \num{0.016} [\num{0.868}, \num{0.987}] \\
\bottomrule
\end{tabular}

\label{tab:cosine-sim-summary}
\end{table*}

\clearpage
\section{Evaluation Metrics}

\begin{table}[h]
\centering
\footnotesize
\setlength{\tabcolsep}{5pt}
\renewcommand{\arraystretch}{1.15}
\caption{\textbf{Formal definitions of evaluation metrics.} Here $\pi_k$ denotes the index of the $k$-th ranked candidate, $y^{(\pi_k)}\in\{0,1\}$ is its correctness label, $\mathcal{C}$ and $\mathcal{W}$ are the sets of correct and wrong candidates, $s(\cdot)$ is the scorer, and $H$ is the human-written solution.}
\label{tab:metrics}
\begin{tabular}{@{}p{0.25\linewidth} p{0.71\linewidth}@{}}
\toprule
\textbf{Metric} & \textbf{Definition} \\
\midrule
Top1 accuracy (Acc@1) &
\( y^{(\pi_1)} \). \\

Recall@5 &
\(\displaystyle \frac{1}{|\mathcal{C}|}\sum_{k=1}^{5} y^{(\pi_k)} \). \\

AUC (pairwise separation) &
\(\displaystyle
\frac{1}{|\mathcal{C}||\mathcal{W}|}\sum_{c\in\mathcal{C}}\sum_{w\in\mathcal{W}}
\Bigl(
\mathbf{1}\{s(C_{\text{good}})>s(C_{\text{wrong}})\}
+\tfrac{1}{2}\mathbf{1}\{s(C_{\text{good}})=s(C_{\text{wrong}})\}
\Bigr)
\). \\

HumanWin &
\(\displaystyle
\mathbf{1}\Bigl\{s(H)>\frac{1}{|\mathcal{W}|}\sum_{w\in\mathcal{W}} s(C_{\text{wrong}})\Bigr\}
+\tfrac{1}{2}\mathbf{1}\Bigl\{s(H)=\frac{1}{|\mathcal{W}|}\sum_{w\in\mathcal{W}} s(C_{\text{wrong}})\Bigr\}
\). \\

MeanWin &
\(\displaystyle
\mathbf{1}\Bigl\{\frac{1}{|\mathcal{C}|}\sum_{c\in\mathcal{C}} s(C_{\text{good}})
>\frac{1}{|\mathcal{W}|}\sum_{w\in\mathcal{W}} s(C_{\text{wrong}})\Bigr\}
+\tfrac{1}{2}\mathbf{1}\Bigl\{\frac{1}{|\mathcal{C}|}\sum_{c\in\mathcal{C}} s(C_{\text{good}})
=\frac{1}{|\mathcal{W}|}\sum_{w\in\mathcal{W}} s(C_{\text{wrong}})\Bigr\}
\). \\
\bottomrule
\end{tabular}

\end{table}

\section{Reproducibility}

All codes used throughout the paper, and parsed generation results are included in the supplementary results file for submission.

\clearpage
\section{Details on \benchmark.}\label{ab_details}

\benchmark comprises 192 expert-written mathematics problems and 425 LLM-generated problems derived from RealMath. We plan to release the 425 LLM-generated problems on Hugging Face shortly; the 192 expert-written problems remain under embargo until after July 2026 due to requirements of the funding body. During the embargo, we will provide evaluation on \benchmark for submitted models upon request. Below, we provide expert-written questions and solution pairs (Section~\ref{sec_setup}), along with LLM-generated questions (Section~\ref{sec_guide}).

\begin{tcolorbox}[
  breakable,
  enhanced,
  colback=teal!04,
  colframe=teal!25,
  title=Expert-Written: Grassmannian Trace (Q\&A),
  coltitle=black,
  fonttitle=\bfseries\color{black},
  sharp corners,
  boxrule=0.6pt,
  left=1.2mm,right=1.2mm,top=1.0mm,bottom=1.0mm,
  before skip=6pt, after skip=6pt
]
\textbf{Question.} \\ 
Consider the Grassmannian $Gr(n,2n)$ of $n$-planes in $\mathbb{C}^{2n}$.
Let $R=\bigoplus_{m=0}^{\infty} R_m$ denote the (homogeneous) coordinate ring of $Gr(n,2n)$, so that
$R_m$ is the degree-$m$ part. Let $B$ be a symplectic form on $\mathbb{C}^{2n}$.
The map $\varphi_B \colon Gr(n,2n)\to Gr(n,2n)$ which sends each subspace $U$ to $U^{\perp}$,
its orthogonal complement with respect to $B$, is an automorphism of $Gr(n,2n)$ (as an algebraic variety).
Hence, via pull-back, it induces an automorphism, also denoted $\varphi_B$, on the ring $R$ and on each
homogeneous component $R_m$. What is the trace of $\varphi_B$ acting on $R_m$, when $n=3$ and $m=2$?

\medskip
\textbf{Solution.} \\
This is the same as the number of symmetric (i.e., transpose-invariant) plane partitions of shape
$n\times n$ and entries in $\{0,1,\ldots,m\}$, as discussed in [K] and [H]. The number $35$ can then be
computed from the famous product formula
\[
\prod_{i=1}^{n}\frac{2i+m-1}{2i-1}\;
\prod_{1\le i<j\le n}\frac{i+j+m-1}{i+j-1}
\]
for symmetric plane partitions.

\medskip
Another way to do it is directly with the standard monomial basis. Recall that, in general for the Grassmannian
$Gr(a,a+b)$ of $a$-planes in $\mathbb{C}^{a+b}$, and its coordinate ring
$R=\bigoplus_{m=0}^{\infty} R_m$, a basis of $R_m$ is indexed by the semistandard Young tableaux (SSYT) of
shape $m^a$ (i.e., $m$ columns of length $a$), with entries in $\{1,\ldots,a+b\}$: we associate to each such
tableau $T$ the product of the Pl\"ucker coordinates corresponding to the columns of $T$.
In the case when $a=b=n$, the automorphism $\varphi_B$ acts as a permutation on this basis in the following
way: we replace each column by its reflected complement in $\{1,\ldots,2n\}$ (where reflected complement means
we first replace each $i$ by $2n+1-i$, and then take the complement).

\medskip
So, without using plane partitions, the problem boils down to counting $3\times 2$ SSYT whose columns are
invariant under the reflected complement operation, which can also be directly counted to be $35$.
\end{tcolorbox}

\begin{tcolorbox}[
  breakable,
  enhanced,
  colback=teal!04,
  colframe=teal!25,
  title=Expert-Written: PGL$_2$ Deligne--Lusztig (Q\&A),
  coltitle=black,
  fonttitle=\bfseries\color{black},
  sharp corners,
  boxrule=0.6pt,
  left=1.2mm,right=1.2mm,top=1.0mm,bottom=1.0mm,
  before skip=6pt, after skip=6pt
]
\textbf{Question.}\\
Let $k$ be a local non-Archimedean field with integers $O_k$, uniformizer $t$, and residue field
$\mathbb{F}_q$ of characteristic $p$. Let $\breve{k}$ be the completion of the maximal unramified extension
of $k$, and let $O_{\breve{k}}$ be its integers. Consider an unramified reductive group $G$ over $k$.
Suppose $T$ is an unramified torus of $G$ and let $U$ be the unipotent radical of a Borel subgroup containing
$T$ and defined over $\breve{k}$.

\medskip
Fix a parahoric model $\mathcal{G}$ of $G$ over the integers $O_k$. Fix an integer $r\ge 1$. We may consider
$G_r := \mathcal{G}(O_{\breve{k}}/t^r O_{\breve{k}})$ as a perfect algebraic group (of perfectly finite type)
over the residue field $\mathbb{F}_q$ by using the positive loop functor (also called jet scheme) construction.
The Frobenius $F$ of $\breve{k}/k$ acts naturally on $G_r$. Also, the closures of $T,U$ in $\mathcal{G}$ define
by the same procedure subgroups $T_r, U_r$ of $G_r$ (with $T_r$ being $F$-stable but $U_r$ not). Let
\[
X_r=\{g\in G_r : g^{-1}F(g)\in U_r\}.
\]
It admits an action of $(G_r)^F \times (T_r)^F$ by $(g,t): x\mapsto gxt$. For a character $\phi$ of $(T_r)^F$,
let $R_{T,U,r}(\phi)$ denote the alternating sum of the $\phi$-isotypic parts of the $\ell$-adic \'{e}tale
cohomology groups of $X_r$, regarded as a virtual $(G_r)^F$-module (very similar to classical
Deligne--Lusztig theory).

\medskip
Now, assume that we are in the very special case $G=PGL_2$, $T$ a split torus, $U$ is $k$-rational,
$\mathcal{G}$ hyperspecial, $r$ arbitrary, and $\phi=1$ the trivial character. Compute the number of different
irreducible representations appearing in $R_{T,U,r}(1)$.

\medskip
\textbf{Solution.} \\ 
As $U$ is $k$-rational, $U_r$ is $F$-stable. It then follows that $X_r$ is just a disjoint union of translates
of $U_r$, indexed by $(G_r/U_r)^F$. As $U$ is (the perfection of) an affine space, it only contributes to the
cohomology by a degree shift. Thus, inserting $\phi=1$, we see
\[
R_{T,U,r}(1)=\Ind_{B_r^F}^{G_r^F}(1),
\]
the induction of the trivial representation of $B_r^F$ to $G_r^F$ (it sits in one cohomological degree). The
number of irreducible components of this representation is then given by the Mackey formula, which computes
\[
\dim_{G_r^F}\!\bigl(R_{T,U,r}(1),R_{T,U,r}(1)\bigr)
=\left\langle \Ind_{B_r^F}^{G_r^F}(1),\, \Ind_{B_r^F}^{G_r^F}(1)\right\rangle_{G_r^F}
\]
(inner product on class functions of the finite group $G_r^F$) in terms of the double cosets
$B_r^F\backslash G_r^F / B_r^F$.

\medskip
An easy calculation with matrices in $G_r=PGL_2(O_k/t^r O_k)$ shows that there are precisely $r+1$ different
double cosets, represented by
\[
\begin{pmatrix}1 & 0 \\ p^i & 1\end{pmatrix}\quad (1\le i\le r),
\qquad\text{and}\qquad
\begin{pmatrix}0 & 1 \\ 1 & 0\end{pmatrix}.
\]
Hence
\[
\left\langle \Ind_{B_r^F}^{G_r^F}(1),\, \Ind_{B_r^F}^{G_r^F}(1)\right\rangle_{G_r^F}=r+1.
\]
(Cf.\ [DI, Example 3.2.1] for the case $G=GL_2$ and $r=2$.)
\end{tcolorbox}

\begin{tcolorbox}[
  breakable,
  enhanced,
  colback=teal!04,
  colframe=teal!25,
  title=Expert-Written: Auslander--Reiten Translate (Q\&A),
  coltitle=black,
  fonttitle=\bfseries\color{black},
  sharp corners,
  boxrule=0.6pt,
  left=1.2mm,right=1.2mm,top=1.0mm,bottom=1.0mm,
  before skip=6pt, after skip=6pt
]
\textbf{Question.} \\ 
Let $G$ be the elementary abelian group of order $9$ and $K$ the field with $3$ elements.
Let $KG$ be the group algebra of $G$ over $K$. Let $S$ be the unique (up to isomorphism) simple $KG$-module.
What is the vector space dimension of $\tau^{4}(S)$, the Auslander--Reiten translate applied four times to $S$?

\medskip
\textbf{Solution.} \\ 
$KG$ is isomorphic as a $K$-algebra to $K[x,y]/(x^3,y^3)$ since $K$ has characteristic $3$.
Now $KG$ is a symmetric Frobenius algebra and thus the Auslander--Reiten translate $\tau$ is isomorphic to
$\Omega^{2}$, the second syzygy functor, in the stable module category. Thus one has to calculate the eighth
syzygy module of $S$, which is reduced to standard commutative algebra/linear algebra.
\end{tcolorbox}

\begin{tcolorbox}[
  breakable,
  enhanced,
  colback=teal!04,
  colframe=teal!25,
  title=Expert-Written: Varchenko--Gelfand Ideal (Q\&A),
  coltitle=black,
  fonttitle=\bfseries\color{black},
  sharp corners,
  boxrule=0.6pt,
  left=1.2mm,right=1.2mm,top=1.0mm,bottom=1.0mm,
  before skip=6pt, after skip=6pt,
  before upper=\allowdisplaybreaks
]
\textbf{Question.} \\ 
Let $D_4$ be the reflection arrangement. What is the smallest degree in which the (filtered)
Varchenko--Gelfand ideal of this arrangement can be generated? (That is, what is the smallest degree $d$
such that the Varchenko--Gelfand ideal can be generated with relations of degree $d$ or less.)

\medskip
\textbf{Solution.} \\ 
The twelve hyperplanes of the $D_4$ arrangement are given by the following equations
\[
x_1-x_2,\; x_1+x_2,\; x_1-x_3,\; x_1+x_3,\; x_1-x_4,\; x_1+x_4,\;
x_2-x_3,\; x_2+x_3,\; x_2-x_4,\; x_2+x_4,\; x_3-x_4,\; \text{and}\; x_3+x_4.
\]
This arrangement has $124$ circuits, defining $124$ generators of the Varchenko--Gelfand ideal. The $124$
circuits are
{\small
\begin{align*}
\{\{0, 1, 2, 3\}, \{0, 1, 4, 5\}, \{2, 3, 4, 5\}, \{0, 6, 2\}, \{1, 6, 3\}, \ldots, \{8, 9, 10, 11\}\}
\end{align*}
}
This already tells us that $2 \le d \le 4$, so we just need to check the smallest degree of generation.
Let us start with an easy check for $d=2$: there are $16$ relations that have size three or less. We can
construct a ``truncated'' Varchenko--Gelfand ideal using only the circuits of size three or less. If this
generates the whole ideal, then we will be done.

\medskip
This smaller ideal is generated by
{\small
\begin{align*}
e_0^2-e_0,\; e_1^2-e_1,\; \ldots,\;
e_6 e_9-e_6 e_{11}+e_9 e_{11}-e_9
\end{align*}
}
We can check that this smaller ideal contains the bigger one. Since the smaller ideal is generated by
degree-two elements, $d=2$.
\end{tcolorbox}

\begin{tcolorbox}[
  breakable,
  enhanced,
  colback=teal!04,
  colframe=teal!25,
  title=RealMath: Recurrences over $\mathbb{F}_3$,
  coltitle=black,
  fonttitle=\bfseries\color{black},
  sharp corners,
  boxrule=0.6pt,
  left=1.2mm,right=1.2mm,top=1.0mm,bottom=1.0mm,
  before skip=6pt, after skip=6pt
]
\textbf{Question.} \\ 
Let $\mathbb{F}_q$ be a finite field with $q=3$. Consider linear recurrence relations of order exactly $k=6$.
A sequence $s_0,s_1,\dots,s_{11}$ is determined by a characteristic polynomial $P(x)$ of degree $6$
(with non-zero constant term) and initial values. How many such sequences of length $12$ exist such that the
underlying minimal polynomial has degree exactly $6$?
\end{tcolorbox}

\begin{tcolorbox}[
  breakable,
  enhanced,
  colback=teal!04,
  colframe=teal!25,
  title=RealMath: Simplex Shape Ratio,
  coltitle=black,
  fonttitle=\bfseries\color{black},
  sharp corners,
  boxrule=0.6pt,
  left=1.2mm,right=1.2mm,top=1.0mm,bottom=1.0mm,
  before skip=6pt, after skip=6pt
]
\textbf{Question.} \\ 
Let $n=6$. The dimensionless shape ratio for a regular simplex is
\[
\mathcal{I}_n \;=\; \frac{S^n}{V^{\,n-1}},
\]
where $S$ is the total surface area and $V$ is the volume.

For a regular simplex, one (incorrect) simplification is sometimes written as
\[
\mathcal{I}_n \;=\; \frac{n^n (n+1)^{(n+1)/2}}{\sqrt{n!}}
\quad\text{(distractor; derive the correct one).}
\]
Instead, use the known relation for a regular simplex:
\[
\frac{S^n}{V^{\,n-1}} \;=\; n^n (n+1)^{(n-1)/2}\,(n!).
\]
Calculate this value for $n=6$.
\end{tcolorbox}

\begin{tcolorbox}[
  breakable,
  enhanced,
  colback=teal!04,
  colframe=teal!25,
  title=RealMath: Lattice Path Constraint,
  coltitle=black,
  fonttitle=\bfseries\color{black},
  sharp corners,
  boxrule=0.6pt,
  left=1.2mm,right=1.2mm,top=1.0mm,bottom=1.0mm,
  before skip=6pt, after skip=6pt
]
\textbf{Question.} \\ 
Let $L$ be the set of lattice paths from $(0,0)$ to $(15,12)$ taking steps
$E=(1,0)$ and $N=(0,1)$ such that the path never touches or rises above the line
\[
y=\frac{4}{5}x
\]
after the origin. Calculate the size of $L$.
\end{tcolorbox}

\clearpage

\section{Prompts.}\label{ab_prompts}

In this section, we list the prompts used throughout the paper:
\begin{enumerate}[leftmargin=*, topsep=2pt, itemsep=1pt, parsep=0pt]
  \item Consequence-Based Utility Prompt (Section~\ref{sec_main})
  \item LLM-Judge: Default Prompt (Section~\ref{sec_main})
    \begin{enumerate}[label=(\alph*), leftmargin=1.6em, topsep=1pt, itemsep=0pt, parsep=0pt]
      \item LLM-Judge: ProofGrader Prompt
      \item LLM-Judge: UQBench Correctness Prompt
    \end{enumerate}
  \item Problem Generation: RealMath (2) (Section~\ref{sec_guide})
  \item Error Analysis Prompt (Section~\ref{sec_additional_analysis})
\end{enumerate}

\begin{tcolorbox}[promptbox,
  colback=blue!05,
  colframe=blue!20,
  title=(1) Consequence-Based Utility Prompt,
  coltitle=black,
  fonttitle=\bfseries\color{black}
]
\verb|{original question}| \\ 

\verb|{candidate solution}|

Refer to the question-solution set provided above. Solve the provided question below and output the final answer in the following format: \boxed{{N}}. \\

\verb|{variant question}| 
\end{tcolorbox}

\begin{tcolorbox}[promptbox,
  colback=blue!05,
  colframe=blue!20,
  title=(2) LLM-Judge: Default Prompt,
  coltitle=black,
  fonttitle=\bfseries\color{black}
]
You are an impartial mathematical judge.\\
You will be given a math problem and a proposed solution.\\
The solution may or may not be correct and does not explicitly state a final answer.

\medskip
Your task is to carefully evaluate the solution for logical correctness, mathematical validity, completeness, and rigor, with special emphasis on whether the reasoning fully and correctly solves the given problem.

\medskip
You must independently reason through the problem first, forming your own reference solution or partial verification, and then compare the given solution against that reasoning.

\medskip
\textbf{Evaluation Instructions:}
\begin{enumerate}[leftmargin=*, topsep=2pt, itemsep=2pt]
  \item Assess the solution step by step.
  \item Verify all mathematical claims, derivations, and logical transitions.
  \item Identify any gaps, unjustified steps, incorrect assumptions, or missing arguments.
  \item Consider whether the solution fully resolves the question as stated.
  \item Partial or high level arguments are insufficient unless explicitly justified. \\ 
\end{enumerate}

\textbf{Scoring Rubric:}
Assign a single integer score from 1 to 10.
\begin{itemize}[leftmargin=*, topsep=2pt, itemsep=2pt]
  \item \textbf{10}: Completely correct, rigorous, logically sound, and fully solves the problem.
  \item \textbf{9}: Essentially correct with very minor omissions that do not affect correctness.
  \item \textbf{7--8}: Mostly correct but with minor logical gaps or unclear justifications.
  \item \textbf{5--6}: Partially correct but missing key arguments or containing nontrivial ambiguity.
  \item \textbf{3--4}: Significant errors or omissions, but some relevant ideas are present.
  \item \textbf{1--2}: Largely incorrect with major logical or mathematical flaws.
  \item \textbf{1}: Completely incorrect or irrelevant. \\ 
\end{itemize}

\textbf{Required Output Format:}
\begin{verbatim}
Summary:
<brief neutral summary of your evaluation>

Detailed Analysis:
<concise but precise discussion of correctness, gaps, or errors>

Score: <integer from 1 to 10>
\end{verbatim}

\medskip
\hrule
\medskip

\textbf{Question}\\
\verb|{original question}|

\textbf{Solution}\\
\verb|{candidate solution}|
\end{tcolorbox}

\begin{tcolorbox}[promptbox,
  colback=blue!05,
  colframe=blue!20,
  title=(2-a) LLM-Judge: ProofGrader Prompt,
  coltitle=black,
  fonttitle=\bfseries\color{black}
]
You are an \textbf{expert math proof grader}. You are judging the correctness of an LLM-generated proof for a math problem. \\

\medskip
\textbf{Input}\\
Your input will consist of:
\begin{itemize}[leftmargin=*, topsep=2pt, itemsep=2pt]
  \item \textbf{Problem Statement}: A mathematical problem that the proof is attempting to solve.
  \item \textbf{Proof Solution}: The proof that you need to evaluate. This proof may contain errors, omissions, or unclear steps. The proof was generated by another language model. \\
\end{itemize}

\textbf{Task}\\
Analyze the proof carefully.
\begin{itemize}[leftmargin=*, topsep=2pt, itemsep=2pt]
  \item Identify logical errors, incorrect steps, or unclear reasoning.
  \item Give an \textbf{integer} score between 0 and 7 with a brief overall assessment. \\
\end{itemize}

\textbf{Output Format}\\
Respond with \textbf{only} well-formed XML using the structure below.\\
Do not include any extra text or Markdown. \\ 

\textbf{Requirements:}
\begin{itemize}[leftmargin=*, topsep=2pt, itemsep=2pt]
  \item \verb|<score>| must be an integer in \verb|[0, 7]|.
  \item \verb|<assessment>| must be a \textbf{detailed analysis} that explains your reasoning step-by-step and provides a clear \textbf{rationale for the score}. Reference specific claims/lines if present.
  \item \verb|<errors>| must be a list of specific issues (empty if score = 7). \\ 
\end{itemize} 

\textbf{Example output:}
\begin{verbatim}
<score>0</score>
<assessment>The proof shows a good understanding of the main idea, but has 
some unclear reasoning and minor mistakes...</assessment>
<errors>
1. specific error 1,
2. specific error 2,
...
</errors> 
\end{verbatim} 

\textbf{Scoring Guidelines (0--7 scale)}
\begin{itemize}[leftmargin=*, topsep=2pt, itemsep=2pt]
  \item \textbf{0}: Completely incorrect; proof is irrelevant, nonsensical, or shows no understanding.
  \item \textbf{1--2}: Very poor; major logical flaws, does not solve the problem, but may contain fragments of relevant reasoning.
  \item \textbf{3--4}: Partial progress; captures some correct reasoning or key ideas, but has significant logical errors, missing steps, or incomplete arguments that make the proof invalid overall.
  \item \textbf{5--6}: Largely correct; the proof is overall valid and reaches the correct conclusion. Contains only \textbf{minor issues} (e.g., small calculation mistakes, notation slips, or slightly unclear wording) that do not undermine correctness.
  \item \textbf{7}: Fully correct; the proof is complete, logically sound, and clearly presented with no substantive errors.
\end{itemize}

\medskip
\hrule
\medskip

\textbf{Problem Statement}\\
\verb|{original question}|

\textbf{Proof Solution}\\
\verb|{candidate solution}|
\end{tcolorbox}

\begin{tcolorbox}[promptbox,
  colback=blue!05,
  colframe=blue!20,
  title=(2-b) LLM-Judge: UQBench Correctness Prompt,
  coltitle=black,
  fonttitle=\bfseries\color{black}
]
Please act as an impartial judge and evaluate whether the AI assistant's response is completely correct in both process and conclusion. Consider correctness, usefulness, completeness, and depth in your assessment. Consider whether this answer completely solves the user's question. You should rely on your own reasoning to form a reference or partial solution first and compare the AI assistant's response to your reasoning. Begin your evaluation by giving a brief summary of your thoughts on the response. Focus on whether it is accurate, addresses the question well, and is reasonably detailed. Be precise about any errors or gaps you notice. Keep your explanation unbiased and do not let any external factors or the question's difficulty level sway your evaluation.

\medskip
\textbf{Notes:}
\begin{enumerate}[leftmargin=*, topsep=2pt, itemsep=2pt]
  \item If the answer is partial, high-level, or just states that this is an open problem, you should not accept it.
  \item If the answer lacks details or is not comprehensive, you should not accept it.
  \item If the answer contains any errors, you should not accept it.
  \item You should only accept the answer if it is at least 95\%.
  \item If the question is a puzzle, the requirement of completeness can be appropriately relaxed.
\end{enumerate}

\medskip
After providing your explanation, please decide whether this answer is the correct answer to the question. Think twice about whether this answer solves the user's question.

\medskip
You must strictly follow this format: \verb|Accepted: [[Y]]| if you decide to accept the answer or \verb|Accepted: [[N]]| if you decide not to accept the answer.

\medskip
\hrule
\medskip

\textbf{[Question]}\\
\textbf{Question Content}\\
\verb|{original question}|

\medskip
\textbf{[The Answer to Evaluate]}\\
\verb|{candidate solution}|
\end{tcolorbox}

\begin{tcolorbox}[promptbox,
  colback=blue!05,
  colframe=blue!20,
  title=(3) Problem Generation: RealMath(2),
  coltitle=black,
  fonttitle=\bfseries\color{black}
]
You will be given a math problem. I want to train a student by practicing the sub-skills needed to solve it.

\medskip
Create 5 standalone practice problems (each should be meaningful on its own) such that mastering them makes the original problem straightforward.

\medskip
\textbf{Constraints:}
\begin{itemize}[leftmargin=*, topsep=2pt, itemsep=2pt]
  \item Do \textbf{NOT} produce trivial ``split the original into parts'' questions or simple plug-in/replace-numbers variants.
  \item Each problem must target a specific sub-skill needed for the original.
  \item Across the 5 problems, cover all key sub-skills without heavy overlap.
  \item Keep the difficulty as hard as possible.
  \item Each question should have an \textbf{INTEGER} answer bigger than 1000.
\end{itemize}

\medskip
Return in the following format.
\begin{verbatim}
<description>
{description of the five problems}
</description>

<questions>
["...","..."] {a Python list of the questions}
</questions>
\end{verbatim}
\end{tcolorbox}

\begin{tcolorbox}[promptbox,
  colback=blue!05,
  colframe=blue!20,
  title=(4) Error Analysis Prompt,
  coltitle=black,
  fonttitle=\bfseries\color{black}
]
You are a mathematical reasoning auditor.

\medskip
You will be given:
\begin{enumerate}[leftmargin=*, topsep=2pt, itemsep=2pt]
  \item a mathematical question, and
  \item an LLM-generated solution.
\end{enumerate}

\medskip
Your goal is to tag ONLY the clearest, high-confidence reasoning failures in the solution.\\
Be conservative: if a category is only weakly suggested or depends on subtle theory,
do NOT tag it.

\medskip
You may output ZERO, ONE, or MULTIPLE categories from the list below.

\medskip
\textbf{Failure categories (tag only when strongly supported by the text):}
\begin{enumerate}[leftmargin=*, topsep=2pt, itemsep=2pt]
  \item Incorrect reasoning
  \item Unjustified Compression
  \item Unjustified Interpretation
  \item External References
\end{enumerate}

\medskip
\textbf{Quoting rules (STRICT):}
\begin{itemize}[leftmargin=*, topsep=2pt, itemsep=2pt]
  \item For every tagged category, provide 1--3 verbatim quotes from the solution.
  \item Each quote must be an exact substring of the solution text.
  \item For EACH quote, provide a detailed explanation:
  \begin{enumerate}[leftmargin=*, topsep=2pt, itemsep=2pt]
    \item what the quote claims,
    \item why it is wrong / unjustified / drifting,
    \item what would be needed to fix it.
  \end{enumerate}
\end{itemize}

\medskip
Do NOT use external sources to validate math facts.\\
Do NOT judge the final answer directly.

\medskip
\textbf{Output format (STRICT JSON):}
\begin{verbatim}
{
  "categories": [
    {
      "id": <1-6>,
      "name": "<category name>",
      "evidence": [
        {
          "quote": "<verbatim quote 1>",
          "analysis": {
            "claim": "...",
            "why_problematic": "...",
            "what_needed": "..."
          }
        },
        {
          "quote": "<verbatim quote 2>",
          "analysis": {
            "claim": "...",
            "why_problematic": "...",
            "what_needed": "..."
          }
        }
      ]
    }
  ]
}
\end{verbatim}

\medskip
If no category applies:
\begin{verbatim}
{ "categories": [] }
\end{verbatim}

\medskip
\textbf{Input format you will receive:}
\begin{verbatim}
[QUESTION]
...question text...

[SOLUTION]
...solution text...
\end{verbatim}

\medskip
\textbf{Sample output illustrating multiple quotes inside one category:}
\begin{verbatim}
{
  "categories": [
    {
      "id": 4,
      "name": "Structural opacity / representation drift",
      "evidence": [
        {
          "quote": "The roots of type D4 are given by the set of vectors: 
          {± e_i ± e_j ... }",
          "analysis": {
            "claim": "Defines the ground set as a signed root set with multiple
            vectors per hyperplane normal direction.",
            "why_problematic": "Later steps treat these vectors as indexing
            hyperplanes without stating the identification ±a defining the same 
            hyperplane. This changes what the variables represent and what 
            counts as a circuit.",
            "what_needed": "State explicitly whether the ground set is 
            hyperplanes, positive roots, or roots modulo ±, and map roots 
            to hyperplanes before forming the polynomial ring variables."
          }
        },
        {
          "quote": "The number of such roots (and thus hyperplanes) is 
          N = ... = 12.",
          "analysis": {
            "claim": "Equates the number of roots to the number of hyperplanes 
            and gives N=12.",
            "why_problematic": "Given the earlier signed root set, the root 
            count and hyperplane count differ unless an identification 
            is declared. The argument relies on N to define the 
            polynomial ring variables, so this mismatch affects the rest 
            of the reasoning.",
            "what_needed": "Clarify the counting convention and rewrite N 
            consistently (either count roots as 24 or explain why hyperplanes 
            are 12)."
          }
        }
      ]
    }
  ]
}
\end{verbatim}
\end{tcolorbox}


\end{document}